%% file: main.tex
\documentclass[10pt,conference]{IEEEtran}
\IEEEoverridecommandlockouts
\usepackage[subfigure]{tocloft}
\usepackage{graphicx}
\usepackage[tight,footnotesize]{subfigure}
\usepackage{hyperref}
\usepackage{wrapfig}
\usepackage{url}
\usepackage{pdfpages}
\usepackage{flushend}
\usepackage{multirow}
\usepackage[titletoc]{appendix}
\usepackage{lipsum}
\usepackage{mdframed}
\usepackage{fancyvrb}
\usepackage{tabu}
\usepackage{tabularx}
\usepackage{multirow}
\usepackage{xspace}
\usepackage{amssymb,amsmath,amsthm}
\usepackage{latexsym}
\usepackage{epsfig}
\usepackage{epstopdf}
\usepackage{xspace}
\usepackage{floatflt}
\usepackage{setspace}
\usepackage{xcolor}
\usepackage{verbatim}
\newcommand{\change}[1]{#1}
\usepackage[compact]{titlesec}
\titlespacing{\paragraph}{0em}{0em}{0em}
\titlespacing{\section}{0em}{0em}{0em}
\titlespacing{\subsection}{0em}{0em}{0em}
\titlespacing{\subsubsection}{0em}{0em}{0em}
\usepackage{soul}
\usepackage[utf8]{inputenc}

\usepackage[sort]{natbib}
\setcitestyle{square}

\usepackage[footnotesize,bf]{caption}
\DeclareCaptionFont{subsecblue}{\color{blue}}
\definecolor{maroon}{cmyk}{0,0.87,0.68,0.32}
\usepackage{enumitem}
\usepackage{color, colortbl}

\def\BibTeX{{\rm B\kern-.05em{\sc i\kern-.025em b}\kern-.08em
    T\kern-.1667em\lower.7ex\hbox{E}\kern-.125emX}}

\begin{document}
\setcitestyle{numbers} 

\setlength{\abovedisplayskip}{0pt}
\setlength{\belowdisplayskip}{0pt}
\title{Comparative Code Structure Analysis using Deep Learning for Performance Prediction}

\author{\IEEEauthorblockN{Tarek Ramadan, Tanzima Z. Islam, Chase Phelps}
\text{\{t\_r297,tanzima,chaseleif\}@txstate.edu}\\
\text{Texas State University}
\and
\IEEEauthorblockN{Nathan Pinnow, Jayaraman J. Thiagarajan}
\text{\{pinnow2,jayaramanthi1\}@llnl.gov}\\
\text{Lawrence Livermore National Laboratory}
}
\maketitle
\vspace{-2in}
\thispagestyle{plain}
\pagestyle{plain}
\vspace{-2in}
\begin{abstract}
\input{abstract}
\end{abstract}


\section{Introduction}
\input{introduction} 
\section{Dataset Description}
\label{sec:dataset}
\input{dataset}

\section{Proposed Approach}
\label{sec:overview}
\input{overview}

\section{Methods}
\label{sec:methodology}
\input{method}

\section{Experimental Setup}
\label{sec:setup}
\input{setup}

\section{Results}
\label{sec:results}
\input{eval}

\section{Discussions}
\label{sec:future-work}
\input{future}

\section{Related Work}
\label{sec:related-work}
\input{related}

\section{Conclusion}
\label{sec:conclusions}
\input{conclusions}

\bibliographystyle{plainnat}
\bibliography{paper}

\end{document}

%% file: abstract.tex
Performance analysis has always been an afterthought during the application development process, focusing on application correctness first. The learning curve of 
the existing static and dynamic analysis tools are steep, which requires understanding low-level details to interpret the findings for actionable optimizations. Additionally, application performance is a function of a number of unknowns stemming from the application-, runtime-, and interactions between the OS and underlying hardware, making it difficult to model using any deep learning technique, especially without a large labeled dataset. In this paper, we address both of these problems by presenting a large corpus of a labeled dataset for the community and take a comparative analysis approach to mitigate all unknowns except their source code differences between different correct implementations of the same problem. We put the power of deep learning to the test for automatically extracting information from the hierarchical structure of abstract syntax trees to represent source code. This paper aims to assess the feasibility of using purely static information (e.g., abstract syntax tree or AST) of applications to predict performance change based on the change in code structure. This research will enable performance-aware application development since every version of the application will continue to contribute to the corpora, which will enhance the performance of the model. \change{We evaluate several deep learning-based representation learning techniques for source code. Our results show that tree-based Long Short-Term Memory (LSTM) models can leverage source code's hierarchical structure to discover latent representations. Specifically, LSTM-based predictive models built using a single problem and a combination of multiple problems can correctly predict if a source code will perform better or worse up to 84\% and 73\%  of the time, respectively.}

%% file: introduction.tex
\begin{table*}
\captionsetup{font=small,skip=0pt}
\centering
\renewcommand{\arraystretch}{1.2}
\begin{tabular}{|l|l|c|c|c|c|c|c|}
\hline
Tag & Contest & Count & Min (ms) & Median (ms) & Max (ms)  & StdDev & Algorithms \\ \hline
A & 4\_C       & 6616  & 86  & 1269   & 4063 & 445 & Hashing~\cite{reg} \\ \hline
B & 230\_B     & 6099  & 31  & 658    & 1872 & 386 & Binary search and number theory~\cite{tprime} \\ \hline
C & 1027\_C    & 832  & 72  & 437    & 1455 & 344  & Greedy~\cite{greedy}\\ \hline
D & 914\_D     & 612   & 206 & 534    & 1965 & 464 & Data structure and number theory~\cite{puzzle} \\ \hline
E & 1004\_C    & 505   & 3  & 80 & 137 & 48 & Constructive algorithm  \\ \hline
F & 1006\_E    & 599   & 51  & 214    & 1647 & 471 & DFS, Graphs, and Trees~\cite{dfs} \\ \hline
G & 1037\_D    & 207   & 5  & 90    & 450 & 63 & DFS, Graphs, and Trees~\cite{dfs} \\ \hline
H & 489\_C    & 5192   & 2  & 9    & 29 & 15 & Dynamic programming (DP) \\ \hline
I & 919\_D    & 475   & 2  & 285    & 800 & 202 & DFS, DP, Graphs \\ \hline
\end{tabular}
\caption{Information about the selected problems. Run times are in milliseconds. The Contest column indicates a specific problem (e.g. $C$ in $4\_C$) in a contest (e.g., $4$ in $4\_C$) from Codeforces. The tags ($A$-$I$) are only assigned for ease of reference.}
 \label{tab:DATA}
 \vspace{-15pt}
\end{table*}

The application development life cycle comprises a perpetual loop of design-development-testing. At least one-fourth of the life cycle for both scientific and commercial applications is spent in rigorous testing. However, performance modeling and analysis have mostly been an afterthought, with program correctness taking center. Most production environments assess program correctness through nightly regression tests to ensure building integrity in commercial and government research laboratories but do not consider performance. 
To combat the absence of performance prediction tools during code development, developers often spend a great deal of time in posterior dynamic (runtime) analysis by executing the target program and measuring the metrics of interest, e.g., time or hardware performance counters. Even though dynamic analysis is more thorough, this process costs hours for data collection and analysis time and requires a human in the loop. 
The cost of dynamic analysis motivates the need for static analysis based on information available before a code runs, such as code structure represented as an abstract syntax tree (AST). \change{This work takes the first step toward developing such static analysis tools that can eventually predict the execution time of applications \textit{before} running them by correlating code patterns to performance problems. While the need for performance regression tests and dynamic analysis will exist, this effort aims at reducing the time and effort spent by developers in dynamic analysis.}

The problem of predicting the absolute execution time of applications based on code structure alone is challenging since the execution time is a function of many factors, including the underlying architecture, the input parameters, and the application's interactions with the OS. Hence, all work in the literature aiming to predict the absolute execution time based on static information (e.g., code structure) alone suffers from poor accuracy~\cite{narayanan2010generating,meng2017mira}. In contrast, this paper circumvents that problem by taking a comparative approach to attribute changes in source code structure to performance $f(\delta_{Code})\to\delta_{Performance}$ and subsequently assess whether a new application will run faster or slower on the same system for similar input. \change{In this context, we define model accuracy as the percentage of times the model (trained on an arbitrary dataset) correctly predicts the label (slower or otherwise) for a test code compared to another one, where the train and the test datasets are disjoint. E.g., $DP$-vs-$DFS$ = 82\% means that a model trained on data from several solutions to a Dynamic Programming $(DP)$ problem can accurately predict the performance difference between a random pair of solutions to the Depth First Search $(DFS)$ problem 82\% of the time.} Since we take a comparative approach, factors that impact applications outside of code structure get nullified. The three use cases of this research are---selecting the best algorithm to solve a problem out of several alternative solutions, predicting performance as a code evolves, and automatically generating semantically similar code suggestions with expected performance change. While the first two use cases can directly apply the source code embeddings generated in this research, the last one requires combining embeddings with code generation. 


To learn the correlation between $\delta(Code)$ and $\delta(Performance)$, we propose a novel static analysis approach that leverages a deep learning technique on Abstract Syntax Trees (AST). 
By directly operating on ASTs, we gain crucial structural information about code while dispensing variations in coding styles. To analyze the AST using deep learning, we build upon techniques initially designed for natural language processing, given the relevant nature between coding language and natural language. We propose to leverage tree-structured Long Short-Term Memory (LSTM)~\cite{tai2015improved}, which automatically learns to represent information inherent to a hierarchical data structure such as an AST to a vector form. 

In this effort, we also present a curated dataset comprising over $4M$ programs, annotated with execution times and memory usage on the same system, from the online programming contest platform \textit{Codeforces}~\cite{codeforces}. We identified thousands of submissions that have significant variations in execution times to build a reliable model. Our work's biggest strength is the models' generalizability since they produce high-quality predictions ($84\%$ accurate on a single problem and $73\%$ on multiple problems) for different algorithms solving different problems. \change{While we do not anticipate entirely dispensing the need for performance analysts during optimizations, our methodology can reduce the amount of time, effort, and resources consumed in a posterior analysis by informing developers of cases where code changes introduce inefficiency.}

In summary, the contributions of this paper include:
\begin{itemize}[leftmargin=*,noitemsep,nolistsep]
\item An extensive labeled dataset of programs representing $1,278$ different problems and $4,313,322$ correct solutions to those problems with varying ranges of execution times. Such a dataset will be useful for further investigating generative deep learning techniques for automatic well-performing code generation.
\item Demonstrate the applicability of deep-learning for static code analysis to learn relationships between code structure changes and performance, assuming code versions run on the same machine.
\item An empirical study to make suitable recommendations for deep learning architecture design, training data sampling, and data augmentation to build predictive models.
\item A pipeline that can be integrated into the development phase of applications to improve prediction accuracy during production.
\item A discussion on the frontiers of generating well-performing code based on the work presented in this paper.
\end{itemize}

%% file: dataset.tex
\begin{figure*}[t]
\captionsetup{font=small,skip=0pt}
  \centering
  \includegraphics[width=\textwidth]{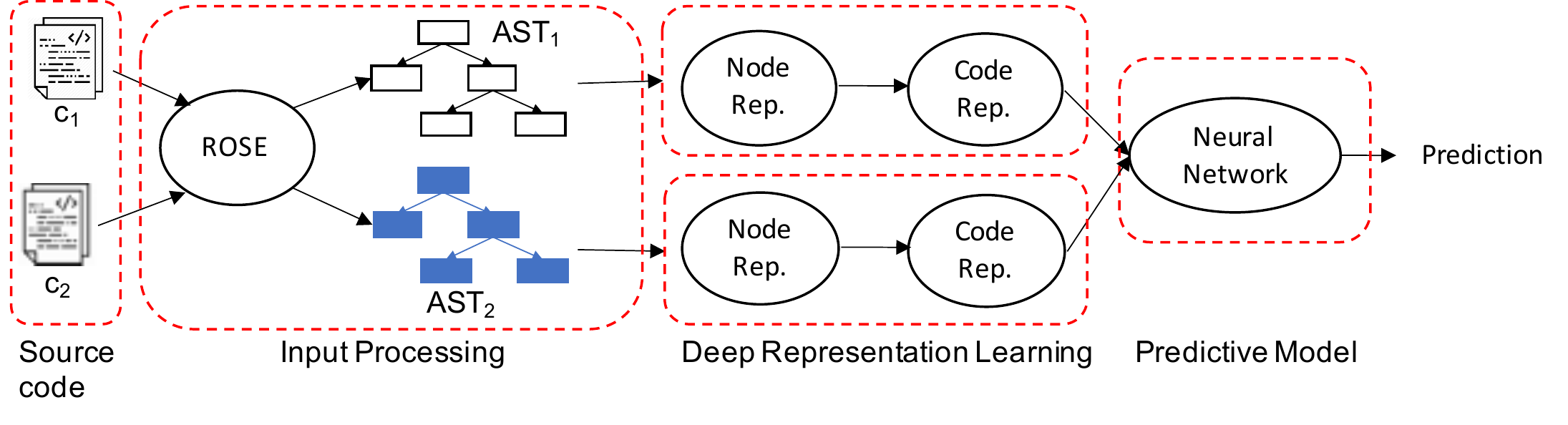}
  \caption{An overview of the proposed deep-learning methodology. Deep representation learning model builds code representation that the predictive model uses to predict if the second code is expected to be more efficient (in terms of execution time).}
  \label{fig:PIPELINE}
\vspace{-0.2in}
\end{figure*}
This section describes the data collection process, along with a brief discussion of the data characteristics. This dataset will be made publicly available via Github along with the pipeline presented in this paper. 

\subsection{Data Collection}
We collected the dataset from an online platform, \textit{Codeforces}~\cite{codeforces}, that organizes programming contests regularly. Each contest consists of a set of problems to which users submit their solutions. The online judge system automatically evaluates each solution for correctness using several test cases (typically 5 to 13, though the number varies among problems) and reports their corresponding runtime and memory usage. The dataset contains several unique solutions to each problem with varying runtime and memory usage characteristics, from which a deep learning model can ``learn". 

We develop a Python tool to automatically retrieve a list of contests from the Codeforces website using their provided API and disregard any contest that has not yet finished. Subsequently, our data collection tool carries out an API request for a list of submission IDs for each contest in the retrieved list. 
For each problem, our tool parses all submissions ignoring those marked by Codeforces as incorrect solutions. 
Finally, our tool enters each problem set along with source code, source language, runtime, and memory usage properties to a database for each test case.

This process results in a total of $4,313,322$ correct solutions, spanning across $1,278$ problems. The distribution of solution times ranges from $6$ problems, each having over $40,000$ submissions to $600$ problems, each having less than $1,000$ submissions. Each problem is unique in regards to its difficulty and popularity. For training, it is crucial to select problem sets that have a sufficient number of solutions and are of sufficient difficulty such that runtime and memory usage across solutions show non-trivial variability. In this paper, we only focus on submissions written in C++, and the tests are averaged to obtain a mean runtime for each problem. However, the tool is generic in its ability to collect all programs written in all languages. We present the performance of the built models in Section~\ref{sec:results} from seven groups of algorithms. Table~\ref{tab:DATA} presents statistics and descriptions of these nine selected problems (Tags A-I). \change{These problems are automatically selected based on sufficient variation in execution times and more than $100$ correct solutions. While not all of the algorithms in our dataset represent scientific applications, Dynamic Programming (DP) and Graph Traversal are two of the $13$ dwarfs of scientific applications, and shortest path algorithms are core to several commercial ones.}

\subsection{Generating Code Pairs}
As described in the previous section, we formulate the problem of comparative performance analysis as correlating $\delta(Code)$ for a pair of source codes with $\delta(performance)$. This formulation takes a differential approach instead of predicting the absolute performance of a new or a variant of the same application, which can be intractable given that execution time is a factor of a large number of variables. 

Our pipeline automatically generates pairs of codes from the selected problems for training a robust model to facilitate this formulation. Since the ordering of every set of codes could be considered as a unique pair, for $N$ submissions, there exists a total of $N^{2}$ possible pairs. Though data-driven approaches such as deep neural networks are typically known to require large amounts of data, we argue that all possible pairs are not required to train a robust model. Not all pairs add unique information for the model to learn, and repetitive training creates a model that has been overfitted. Hence, in Section~\ref{sec:dataneed}, we evaluate the data requirements to build a robust model. The state-of-practice method for reducing the training dataset is to use random subsets of input. Hence, we explore the use of random subsets (of code pairs) of varying sizes for training the models and make appropriate recommendations. For every pair of programs, we generate the target variable (binary) as follows: if the first element of the pair has a higher execution time, we label it as positive, otherwise negative. This formulation emulates a developer looking to determine if a new version of the program will have a lower (improved) runtime. Our experiments also study the impact of including two-way ordering of every pair of codes, i.e., $(a, b)$ and $(b,a)$ on the model accuracy.

%% file: overview.tex
  
  
\begin{figure*}[t]
\captionsetup{font=small,skip=0pt}
\centering
  \includegraphics[width=\textwidth]{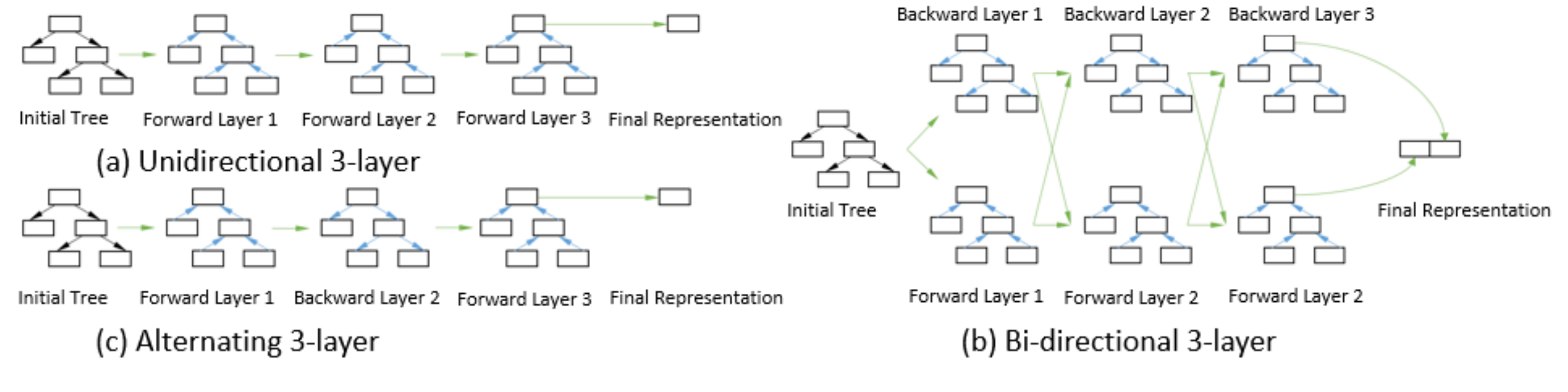}
\caption{Proposed training strategies based on tree-structured LSTMs for processing ASTs. Arrows going from root to leaves (and vice versa) indicate information flow in the LSTM, arrows between trees indicate information flow between layers. 
}
  \label{fig:3LAYER}
  \vspace{-0.2in}
\end{figure*}

Figure~\ref{fig:PIPELINE} shows an overview of our approach that converts each source code to an AST by using the ROSE~\cite{dan2011rose} compiler and subsequently utilizes a deep neural network to learn embeddings from ASTs that incorporate structural information automatically.
Embedding learning is an active area of research with several algorithms such as deep graph convolutional network~\cite{schlichtkrull2018modeling}, deep graph attention network~\cite{wang2019heterogeneous}, and LSTM~\cite{sundermeyer2012lstm}. 
An adequate representation improves the accuracy of the downstream analysis tasks (e.g., performance prediction in our case) by capturing discriminatory information that contributes to determining a label. 
These embeddings then become the input to a classifier module that implements a fully connected neural network to perform the prediction. While this paper mainly focuses on ASTs to build a deep learning pipeline for static program modeling, we expect that information gathered during compile-time, such as control and data flow graphs, may improve a model's accuracy. 

\subsection{Problem Formulation:}
Formally, let us denote the ASTs for a pair of source code by $p_i$ and $p_j$ respectively, where every $p_i \in \mathrm{P}$ is a submission pertinent to the problem $\mathrm{P}$. We define a deep feature extractor $F$ that processes an AST to produce a latent representation ${z} \in \mathrm{Z}$, where $\mathrm{Z}$ denotes the latent space. Mathematically, this process is expressed as $F: \mathrm{P} \mapsto \mathrm{Z}$. Since the goal is to predict if the AST of $p_j$ is expected to have a lower execution time than $p_i$, we first concatenate their features to produce $\bar{{z}}_{ij} = [{z}_i, {z}_j]$. As a result, when the dimensionality of the latent space $\mathrm{Z}$ is $d$, the size of the concatenated feature $\bar{{z}}_{ij}$ becomes $2*d$. The classifier function $C$ maps the concatenated feature $\bar{{z}}_{ij} \in \bar{\mathrm{Z}}$ into the target variable $y_{ij} \in \mathrm{Y}$, where the output space $\mathrm{Y}$ is discrete and assumes one of the two values $0$ or $1$. In other words, the classifier mapping can be expressed as $C: \bar{\mathrm{Z}} \mapsto \mathrm{Y}$.

\change{The feature extractor $F$ consists of two components. First, it learns to represent each code construct of an AST (\textit{node} of a tree) using an embedding lookup function that assigns a feature vector to each node depending on its type (e.g., \texttt{for loops} or \texttt{if statements}). Second, it learns representation for the entire AST or a sub-tree using the deep learning algorithm ($\mathbf{z}$). In this paper, we propose tree-LSTM for automatic feature representation. Section~\ref{sec:tree-lstm} discusses our rationale in details. Our proposed approach jointly infers the representations for each of the nodes and subsequently for the entire AST. This automated learning approach alleviates the need for any manual feature engineering technique, which presents a nontrivial challenge for source code.} The classifier is implemented as a feed-forward network and produces the likelihood of one version of the code being superior to the other in terms of expected performance. This likelihood is then encoded into a decision label using Equation~\ref{eqn:decision}:

\begin{gather}
pred(p_{i}, p_{j}) \rightarrow 
\begin{cases}
  0 & t_{i} < t_{j} \text{; } p_{i} \text{is faster}\\
  1 & t_{i} \ge t_{j} \text{; } p_{j} \text{is faster or equivalent}
\end{cases}
\vspace{-0.2in}
\end{gather}

\subsection{Tree-Structured LSTMs for AST Modeling}
\label{sec:tree-lstm}

Since conventional solutions such as fully connected networks and convolutional neural networks typically utilize unstructured high-dimensional and image data, they can not directly handle ASTs. 
Consequently, there has been a recent surge in deep learning techniques designed specifically for these challenging data types. Tree LSTM~\cite{tai2015improved} is a deep learning architecture for processing tree-structured data.
Recent work has demonstrated that leveraging the hierarchical structures of languages, both natural and programming, gives models salient characteristics of the data and improves performance on the downstream tasks~\cite{tai2015improved,eriguchi2016tree}.
Consequently, we propose using tree-structured LSTM, a specific construction of recurrent neural networks, to produce concise vector representations for each source code. 

Intuitively, our model uses hierarchical accumulation to encode each non-terminal node's representation by aggregating the hidden states of all of its descendants. The accumulation process occurs in two stages. First, the model induces the value states of non-terminals with hierarchical embeddings, which helps the model become aware of the hierarchical and sibling relationships between the nodes. Second, the model performs an upward cumulative-sum operation on each \textit{target node} \change{(the node whose representation is being learned)}, accumulating all elements in the branches originating from the target node to its descendant leaves. In Section~\ref{sec:results}, we thoroughly evaluate the performance of tree-LSTM with Graph Convolution Network (GCN) based deep representation learning models suitable for structured graph data. \change{We demonstrate that the accuracy of the predictions are higher with embeddings built using tree-LSTM compared to that of GCN.} 


Broadly, tree-LSTM is a recurrent neural network (RNN)~\cite{williams1989learning}, designed to perform feature extraction from arbitrary length sequence data via the recursive application of a transition function on a hidden state vector. It operates by taking in the hidden state of the previous element of the sequence and the input for its current element. At each time step $t$, the hidden state vector $h_{t}$ is a function of the input vector $x_{t}$ (vector representation of the $t^{th}$ element in the sequence) and its previous hidden state $h_{t-1}$. Consequently, the $h_{t}$ can be interpreted as a concise representation of the sequence of elements observed up to time $t$. In a typical RNN, the transition function is implemented as follows:
\begin{equation}
    h_t = \tanh{W x_t + U h_{t-1} + b},
\end{equation}\noindent
where $W$, $U$, and $b$ are learnable parameters, and $\tanh$ denotes the hyperbolic tangent nonlinearity. An inherent limitation of RNNs is that as the sequence length grows the problem of \textit{exploding} or \textit{vanishing} gradients makes training very difficult~\cite{pascanu2013difficulty}. Consequently, the LSTM architecture addresses this limitation in learning long-term dependencies by introducing a memory cell that preserves the states over long periods of time~\cite{sundermeyer2012lstm, graves2005framewise}. For a time step $t$, an LSTM unit is typically comprised of vectors from an input gate $i_{t}$, forget gate $f_{t}$, output gate $o_{t}$, memory cell state $c_{t}$ and hidden state $h_{t}$. Intuitively, the forget gate controls the extent to which the memory cell's previous states are forgotten, the input gate controls how much each unit is updated, and the output gate controls the exposure of the internal memory state. In a nutshell, the hidden state vector in an LSTM unit is a gated, partial view of the internal memory cell state. Mathematically, the transition equations can be derived as follows:
\begin{align}
i_t &= \sigma(W^i x_t + U^i h_{t-1} + b^i),\nonumber \\
f_t &= \sigma(W^f x_t + U^f h_{t-1} + b^f),\nonumber  \\
o_t &= \sigma(W^o x_t + U^o h_{t-1} + b^o),\nonumber  \\
u_t &= \sigma(W^u x_t + U^u h_{t-1} + b^u),\nonumber  \\
c_t &= i_t \odot u_t + f_t \odot c_{t-1},\nonumber  \\
h_t &= o_t \odot \tanh({c_t}).
\end{align}

A limitation of this architecture is that it allows only strictly sequential information propagation. However, in the case of ASTs, the information flow happens to multiple children from a given parent node. \change{Hence, we propose a new architecture for the tree-LSTM technique to deal with information flow through an AST~\cite{tai2015improved}}. The crucial difference between an LSTM unit and a tree-LSTM unit is that the gating vectors and memory cell updates are dependent on the states of possibly many child units. Additionally, instead of a single forget gate, the tree-LSTM unit contains one forget gate for each child, which allows it to selectively leverage information from each child. Section~\ref{sec:NODEREP} describes the input vector at each node. Figure~\ref{fig:3LAYER}(a) shows that the transition function is applied to the leaf nodes first and then progressively moves up the tree to the root node. Mathematically, this can be described as follows: Assuming that the $\mathcal{C}(j)$ denotes the set of children of a node $j$,
\begin{align}
\tilde{h}_j &= \sum_{k \in \mathcal{C}(j)} h_k,\nonumber  \\
i_j &= \sigma(W^i x_j + U^i \tilde{h}_j + b^i),\nonumber \\
f_{jk} &= \sigma(W^f x_j + U^f h_k + b^f), \nonumber \\
o_j &= \sigma(W^o x_j + U^o \tilde{h}_j + b^o), \nonumber \\
u_j &= \sigma(W^u x_j + U^u \tilde{h}_j + b^u), \nonumber \\
c_j &= i_j \odot u_j + \sum_{k \in \mathcal{C}(j)} f_{jk} \odot c_k,\nonumber  \\
h_j &= o_j \odot \tanh({c_j}).
\end{align}Note that, after processing the entire AST \change{(or a selected sub-tree)}, the classifier function uses the final hidden representation at the root node (of the sub-tree) for prediction.

%% file: method.tex
As described in the previous section, our goal is to jointly infer node representations (depending on the node type) and AST representations, with the overall objective of predicting performance. To achieve this goal, we need to address the following tasks: (1) generate ASTs, (2) represent nodes of an AST as a vector (encoding lookup), and (3) combine node embeddings for representing source code (AST modeling).
In this section, we describe how we address each of the tasks in detail, along with implementation specifics. We follow the notations introduced in the previous section. \subsection{AST Generation}
\label{sec:GENAST}
The first step towards applying deep neural networks to codes is to create an appropriate representation. In general, the code can be treated as a text excerpt and processed with standard language modeling tools such as word or document embeddings. However, we advocate using abstract syntax trees since they are better descriptors of a code structure. This transformation leads to an additional challenge that the neural network should leverage the inherent tree structure of ASTs.

To generate the ASTs, we use the ROSE~\cite{dan2011rose} compiler infrastructure. ROSE is a flexible, portable, and scalable source-to-source compiler infrastructure widely used in the scientific community comprising the national laboratories, universities, and the industry. \change{The AST from ROSE is modified only to include internal nodes that are part of the source code's function definitions. This process removes irrelevant information from the tree and allows models to train faster.} For simplicity, the source code's function definitions are all set as children of a root node. \change{While these simplifications make the embedding learning process simpler, more fine-grained representation of the tree nodes could provide additional information for the model to exploit. Finally, the AST generation process outputs a list of the node IDs and a list of links between nodes to represent the tree.}

\subsection{Constructing Node Embeddings}
\label{sec:NODEREP}
We assign a unique ID to each type of internal node (e.g., \texttt{for}, \texttt{while}), consistent across all trees in the database. A node type gets the same ID even when they appear multiple times in the same tree. Following standard practices in the natural language processing area, our machine learning pipeline transforms the user-defined tokens (IDs in ASTs) into vector representations. A naive approach for constructing a vector representation from an AST is using one-hot encoding, where each ID can be assigned a $1-$sparse binary vector with the value $1$ at the location of the ID and $0$ elsewhere. Since such a vector's size is a function of the total number of unique IDs in the dataset, it is high-dimensional (often referred to as \textit{cursed} representations). Such representations can lead to severe overfitting when building predictive models. 

Hence, we investigate a different approach in this work that assigns a specific vector representation for each ID using an embedding lookup structure. We fix the dimensionality of the embedding at $\lambda$, and initialize the embeddings randomly. However, the neural network training process can subsequently tune the embeddings. The total number of parameters to be optimized in this step is $\lambda \times D$, where $D$ is the total number of unique IDs in our dataset. This embedding layer will allow the model to infer similarities between nodes in terms of their performance impact, much more effectively than simple one-hot encoding. Once encoded, these embeddings are passed along 
to train a model for generating representations for the entire tree. In this paper, we initialize using random embeddings. In the future, we will investigate using pre-trained embeddings by adapting word embedding techniques (e.g., Skip-gram~\cite{mikolov2013efficient}, GloVe~\cite{pennington2014glove}) from the Natural Language Processing (NLP) literature.

\subsection{Training Models}
Following the state-of-practice in neural networks, we advocate using multiple layers in our tree-LSTM architecture. In this design, the hidden states at the end of one layer are used as the next layer's node representations. This process typically leads to greater refinement of each sub-tree's representations as each layer provides a better representation of the tree structure. In addition to this native implementation, which we refer to as \textit{uni-directional} tree-LSTM, we also consider two variants:  \textit{bi-directional} tree-LSTM, and \textit{alternating} tree-LSTM. In the first variant, we allowed the tree transition to be bi-directional, i.e., from root to leaf nodes and leaf to root nodes. As illustrated in Figure \ref{fig:3LAYER}(b), two different tree-LSTMs run independently, with one having hidden states going from child to parent and the other going from parent to child. The parent node copies its representation to all its children instead of just sending its representation to a single node. This information propagation pattern enables a node's hidden state to include information from its children and its ancestors. Finally, the two representations are concatenated together to form a unified representation for a node. Since our approach uses the final root node representation to make the prediction, the downward pass in the bi-directional training's final layer is not required. In the second variant of the tree-LSTM, bi-directional training is simplified by alternating forward and backward passes. As shown in Figure \ref{fig:3LAYER}(c), a $3-$layer tree-LSTM will be constructed with $2$ forward layers and $1$ backward layer between them. Compared to the bi-directional training, this contains only half the number of parameters to train, avoids overfitting in practice, and produces highly effective latent representations for ASTs. In our experiments, we find that the alternating tree-LSTM consistently produces the best performance in all cases. 

\subsection{Classifier Design}
\label{sec:classifier}
Since our approach's overall objective is to perform comparative analysis, we first concatenate the hidden representations for the two ASTs and subsequently pass it to a fully connected classifier with the sigmoid activation. This classifier's number of parameters is $2*d$, where $d$ is the size of the latent representation in our tree-LSTM. We then compute the binary cross-entropy loss between the predicted probabilities and the correct labels to optimize the network's unknown parameters. 


%% file: setup.tex
This paper evaluates our proposed methodology using nine individual problems (presented in Table~\ref{tab:DATA}), and a combined dataset comprising 100 submissions picked randomly from 100 different problems (referred to as \textit{MP} in Figure~\ref{fig:boxplot}). \change{We calculate the accuracy of a model based on the percentage of times the model correctly classifies a code as slower or otherwise compared to another one.} 
\subsection{System}

We run our experiments on the Google Cloud Platform (GCP). 
Specifically, we use a machine with eight virtual CPUs, 30 GB RAM, and one NVIDIA Tesla P100 GPU with 16GB memory. Each CPU consists of 2.30GHz Intel (R) CPU with 4 cores and is based on the $x86\_64$ Architecture. The GPUs are equipped with CUDA 10.0 toolkit by NVIDIA.
\subsection{Deep Representation Learning Techniques}
\label{sec:gcn}
In this paper, we evaluate the efficacy of our proposed tree-LSTM (described in Section~\ref{sec:tree-lstm}) based representation learning technique compared to that of a more generic one---Graph Convolution Network (GCN)~\cite{kipf2016semi,schlichtkrull2018modeling}. The GCN is a generalization of Convolution Neural Networks (CNN)~\cite{kalchbrenner2014convolutional}, where the GCN stacks multiple graph convolution layers to extract a high-level node representation that takes $N$-dimensional data to graph data. A GCN model takes the graph-structured data as input and generates a vector representing a source-code. 
The GCN applies semi-supervised node classification, which classifies each node of the tree to help decide the type for the whole AST.
We extend the GCN model by creating a wrapper layer that combines information from an internal node's directly connected nodes. The significant difference between GCN and tree-LSTM is in information flow to each internal node--GCN leverages all neighboring nodes compared to tree-LSTM which leverages knowledge of parent-child relationships. The source-code embeddings are then passed from GCN to the classifier, as described in Section~\ref{sec:classifier}.

\subsection{Hyper-parameter Tuning}
\label{sec:hyper}
For automated hyper-parameter tuning, we leverage the Optuna optimization framework~\cite{akiba2019optuna}. For GCN, we observe that the most critical parameters to tune for the given downstream prediction task are the number of convolution layers and the hidden layer's size. We vary the number of convolutional layers from 1 to 16 for GCN (more than that exhausts GPU memory) and the hidden layer's size from 8 to 256. Our experiments show that (6, 117) as the number of convolutional layers and the hidden layer's size, respectively achieves the best accuracy ($68.5\%$). For tree-LSTM, we use $100$ hidden states and the feature embedding vectors of length $120$.
With these parameters, we achieve the best accuracy ($73\%$) for the large dataset.

%% file: eval.tex
In this section, we present a detailed evaluation of the proposed approach. The details about the dataset can be found in  Section~\ref{sec:dataset}. We used a subset of the submissions (disjoint from training dataset) to demonstrate results. In this study, we evaluate the following: (a) the effectiveness of our proposed tree-LSTM architecture compared to GCN in building source-code representations and the generalizability of the predictive models built on them; (b) the architectural choices for the best model; (c) the design choices for data sampling and augmentation; and (d) the prediction \change{sensitivity} to runtime variation.

\subsection{Model Evaluation and Generalization}
\label{sec:LAYERS}
\begin{figure}[t]
\vspace{-0.1in}
\captionsetup{font=small,skip=0pt}
  \centering
  \includegraphics[width=\columnwidth]{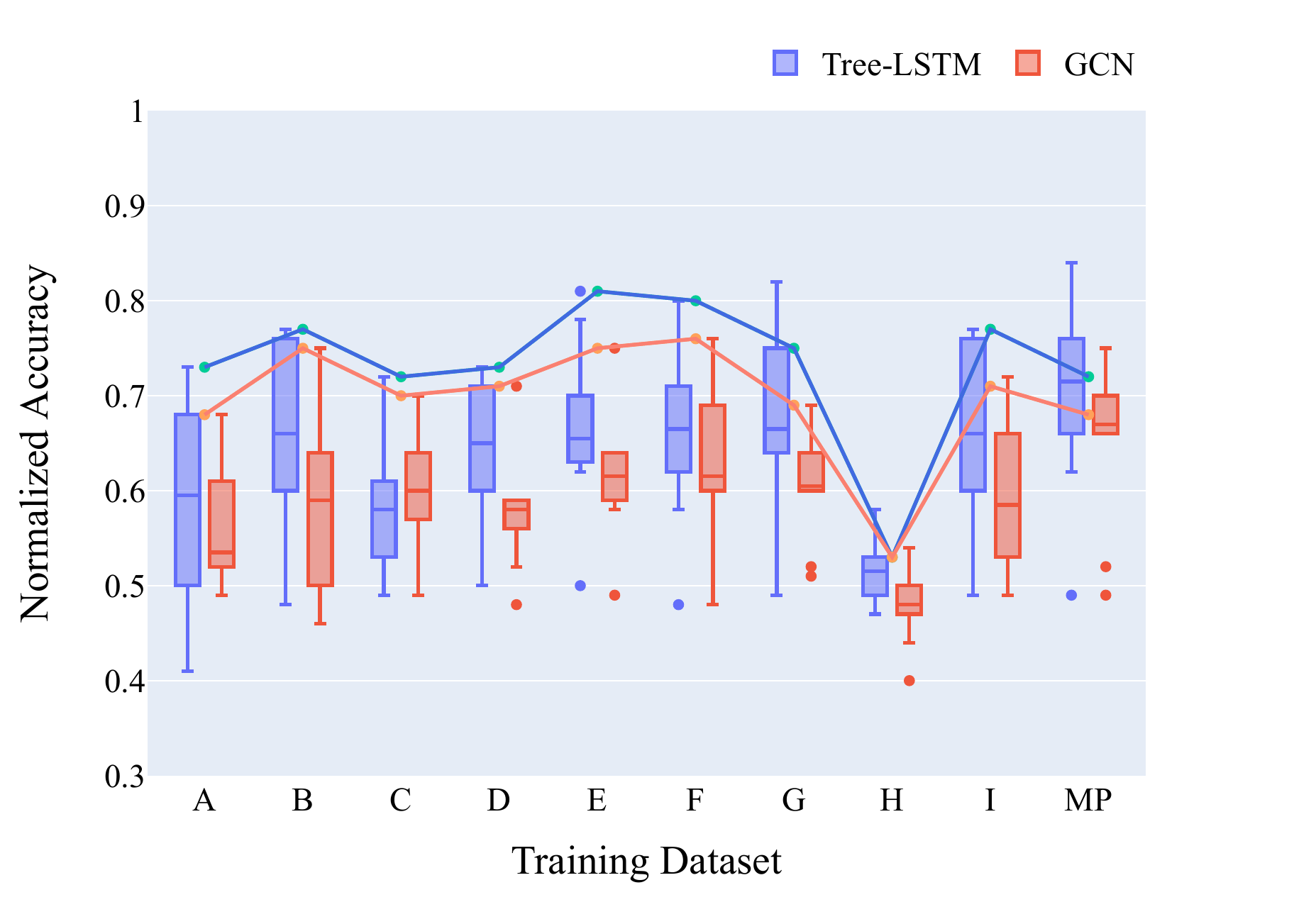}
  \caption{The overall model evaluation and generalizability of our proposed tree-LSTM approach compared to a traditional GCN method. The X-axis shows the training dataset, and the Y-axis shows the accuracy. The lines show the accuracy of models in classifying the performance differences (lower or otherwise) between random pairs of disjoint submissions for the same problem (as the training problem). The boxplots show models' accuracy in classifying the performance difference between random pairs of disjoint submissions from all others problems except the training one.}
  \label{fig:boxplot}
  \vspace{-0.1in}
\end{figure}

This experiment's objective is two-fold--(a) test the effectiveness of our proposed tree-LSTM architecture compared to GCN in building source-code representations and their impact on model accuracy, and (b) determine how well the predictive model can generalize to unseen problems. For this experiment, we train models on submissions from a problem and measure the accuracy of the model in predicting the change in performance (positive or negative) for unseen submissions from (i) the same problem \change{(disjoint set)}, (ii) different problem but the same algorithmic group, and (iii) other problems from diverse algorithmic groups. \change{Figure~\ref{fig:boxplot} shows training datasets along the X-axis. The boxplots along the Y-axis show models' accuracy in classifying the performance difference between random pairs of disjoint submissions from diverse problems. The line plots show the accuracy of $p_{i}$ vs $p_{i}$, where the training and testing datasets are disjoint submissions of the problem $p_{i}$.}

\noindent \textbf{Generalization: }
\change{From Figure~\ref{fig:boxplot}, we can observe that a model built for a specific problem predicts the label of a disjoint set of submissions from (i) the same problem with up to 81\% accuracy (line chart for training set \texttt{E}); (ii) different problems from different algorithmic groups with up to 80\% accuracy (boxplot for the problem \texttt{E}). Further investigation shows that the highest accuracy is incurred when the constructive algorithm to solve problem \texttt{E} classifies random pairs of submissions from a DFS problem \texttt{G}. 
Also, a model built on the same problem can accurately classify the difference in execution times between random pairs of disjoint submissions from other problems from the same algorithmic group with up to 82\% accuracy. E.g., model built on DFS problem \texttt{F} classifies submissions from another DFS-based problem \texttt{G} with up to 82\% accuracy. These observations show that the model is learning problem characteristics and not just remembering programming constructs.}

\change{To test our approach's generalizability, we build a model using a large dataset by randomly selecting 100 submissions from 100 different problems with sufficient variation in execution times and more than 1000 correct solutions. We denote this dataset as \texttt{MP} (for multiple problems). We then evaluate a model trained on \texttt{MP} to predict performance differences of submissions from both A-I and these $100$ problems (disjoint test set). Figure~\ref{fig:boxplot} shows that the predictive model trained on \texttt{MP} can accurately classify performance difference between random pairs of solutions with up to 84\% accuracy for A-I (boxplot) and 73\% accuracy for a disjoint set of \texttt{MP} submissions (line chart).}

\noindent \textbf{Tree-LSTM vs GCN: }
From Figure~\ref{fig:boxplot} we can observe that the prediction task with the tree-LSTM-based embeddings consistently outperforms that built using the GCN model. 
The tree-LSTM based representations capture crucial hierarchical information about code structure that a generic graph-based model fails to do, which explains GCN's poor performance.
\begin{table}[t]
\captionsetup{font=small,skip=0pt}
\renewcommand{\arraystretch}{1.2}
\centering
\begin{tabular}{|c|c|c|c|}
\hline
    & F    & G    & I    \\ \hline
F   & .80 & .72 & .67 \\ \hline
G   & .82 & .76 & .68 \\ \hline
I   & .76 & .67 & .77 \\ \hline
\end{tabular}
\caption{Models trained and evaluated on different problems in similar algorithm groups (DFS and Graphs). Rows indicate the training dataset, and columns display the test set. While problems F and G share the same algorithmic classes (DFS, Graphs, and Trees), the problem I has a partial overlap (DFS, DP, Graphs). This result indicates that a more considerable overlap in problem characteristics will result in higher prediction accuracy.}
\vspace{-10pt}
\label{table:GEN}
\end{table}

\begin{table}[t]
\footnotesize
\renewcommand{\arraystretch}{1.2}
\centering
\begin{tabular}{|c|c|c|}
\hline
\rowcolor{gray!20}
Layers & Uni-Directional & Bi-Directional \\ \hline
\multirow{2}{*}{1} 
    &0.773 & 0.769\\\cline{2-3}
    &0.780 & 0.78 \\ \hline

\multirow{2}{*}{2} 
    &0.765 & 0.767\\\cline{2-3}
    &0.789 & 0.786 \\ \hline
    
\multirow{2}{*}{3} 
    &0.766 & 0.77\\\cline{2-3}
    &0.783 & 0.767 \\ \hline \hline
\rowcolor{gray!20}
\multicolumn{3}{|c|}{Alternating layers}   \\ \hline
\multicolumn{3}{|c|}{\textbf{0.77} (A) and \textbf{0.804} (C)}                 \\ \hline
\end{tabular}
\captionsetup{font=small,skip=0pt}
\caption{Prediction performance of the proposed approach \textit{problem set A and C}. We report the results obtained using different architectural choices.}
\label{table:LAYERS}
\vspace{-0.2in}
\end{table}
\begin{figure}[t]
\captionsetup{font=small,skip=0pt}
  \centering
  \includegraphics[width=.4\textwidth]{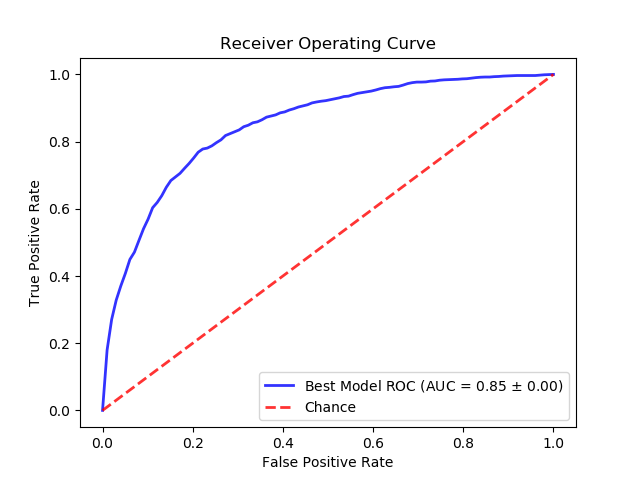}
  \caption{ROC curve on the validation set obtained using the multi-layer alternating Tree-LSTM architecture on \textit{problem set A}.}
  \label{fig:ROC}
  \vspace{-0.2in}
\end{figure}

\subsection{ROC vs Accuracy Metric}
In addition to the accuracy metric, we evaluate all our models based on the receiver operator curve (ROC) to study how the prediction task's performance varies as the confidence threshold changes. The confidence threshold of the models' output (probability) determines if the performance difference between two code pairs should be classified as positive or negative. Increasing the confidence threshold lowers the false positive rate and equivalently the true positive rate. Having a lower false positive means that if a model classifies the change in execution time as increasing, the application developer can confidently invest the time and effort to resolve coding inefficiencies (perhaps by using other tools).
E.g., in Figure \ref{fig:ROC}, we can observe that the ROC for the problem \texttt{A}, obtained using the $3-$ layer alternating tree-LSTM architecture, achieves a high area under the ROC metric of $0.85$, which implies $85\%$ true positive rate. Since this measure agrees with the accuracy metric, we only report the experiments' accuracy scores.

\subsection{Impact of Architectural Choice}
This experiment's objective is to evaluate the impact of the different architectural choices for the tree-LSTM (the best representation learning model for source-code, as found in Section~\ref{sec:LAYERS}) on the overall prediction accuracy. 
Table~\ref{table:LAYERS} presents the impact of three architectural choices on the accuracy of the prediction task. We increase the number of layers from $1$ to $3$ for the uni- and bi-directional architectures and observe an insignificant change in the accuracy. The bi-directional architecture is significantly more complicated and takes much longer to train since information is combined from both the forward and backward passes at every layer. The lack of improvement in model accuracy indicates overfitting due to the arbitrary increase in model complexity. The alternating architecture produces an equivalent representation compared to the other architectures, e.g., the alternating architecture improves the downstream predictive task's performance by 2\% for problem C. However, the alternating architecture combines information once during the forward pass followed by the backward pass, thereby gathering more information than a uni-directional one. In comparison, its accuracy is similar to that of the bi-directional architecture while being faster to train.

\subsection{Impact of Data Sampling and Augmentation}
\label{sec:dataneed}
\begin{figure}[t]
\captionsetup{font=small,skip=0pt}
  \centering
  \includegraphics[width=\columnwidth]{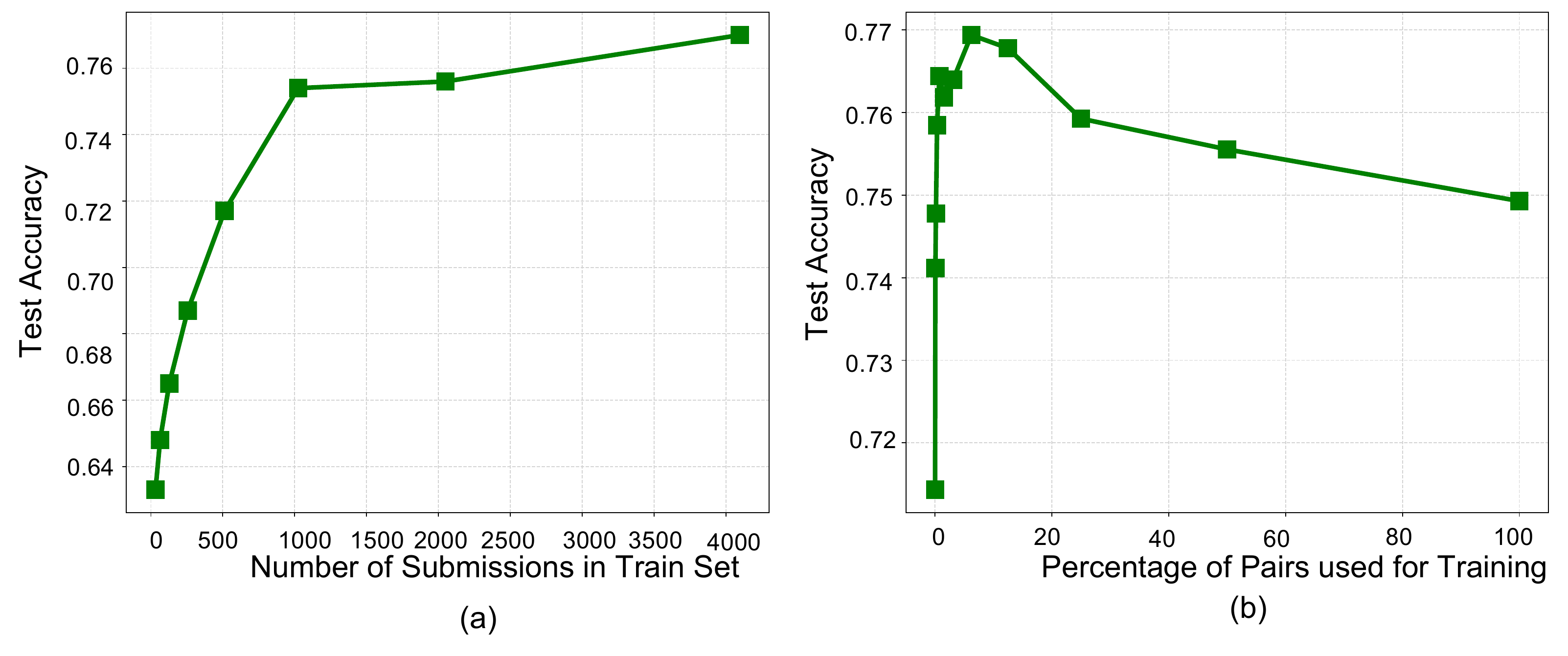}
\caption{(a) Accuracy of the model based on a percentage of maximum pairs for 2048 submissions. (b) Accuracy changes based on training set characteristics. }
\label{fig:pairs}
\vspace{-0.2in}
\end{figure}
A crucial aspect of modern machine learning methods is the complex trade-off between the task complexity and the amount of training data required to produce reliable models. Hence, we study the impact of data sampling on our approach's observed performance. We vary the number of submissions, with a fixed ratio of pairs for every case and the number of pairs for a given number of submissions. 

\noindent \textbf{Impact of number of submissions during training: }
First, we increase the number of submissions in the training set from $32$ to $4096$ by powers of $2$. For each case, we construct the training pairs by selecting a random $75\%$ of all possible pairs in that case. For all tests, we use the same test set of submissions. Figure \ref{fig:pairs}(a) shows results for the problem set A with the multi-layer alternating tree-LSTM architecture. Figure \ref{fig:pairs}(a) shows that the accuracy steadily improves as the number of submissions grows. However, beyond $1000$ submissions, there is a diminishing return. Since data collection and annotation are time-consuming, having a reasonable number of training samples improves this methodology's adaptability in practice.

\noindent \textbf{Impact of the percentage of pairs during training: }
We also investigate the question of how many pairs should be included from those submissions for training. We perform this study by increasing the percentage of pairs used for a fixed number of submissions. In particular, we select the number of submissions at $2048$ and vary the ratio of pairs used. Interestingly, Figure \ref{fig:pairs}(b) shows that the accuracy initially improves rapidly as the number of pairs (randomly chosen) increases, achieving an accuracy improvement of 10\%, however, the accuracy score then begins to dip. However, there is a dip in the accuracy score. We continue to include more pairs since complex models such as deep networks tend to overfit when the training data's complexity is high. This observation motivates the need for further investigation of sampling strategies for optimal performance.

\noindent \textbf{Impact of the ordering of pairs during training: }
We evaluate how important it is for the model to train on both orderings of a single pair ($(a,b)$ and $(b, a)$). We compare a training set containing only one ordering pair to a model trained on the same number of overall teams, with half being the others' reverse. We find that the accuracy improves marginally, up to 2\% from using symmetrical pairs as opposed to non-symmetrical ones (the figure is not included due to space limitation).

\subsection{Prediction Sensitivity}
\label{sec:sensitivity}
\begin{figure}[t]
\captionsetup{font=small,skip=0pt}
  \centering
  \includegraphics[width=\columnwidth]{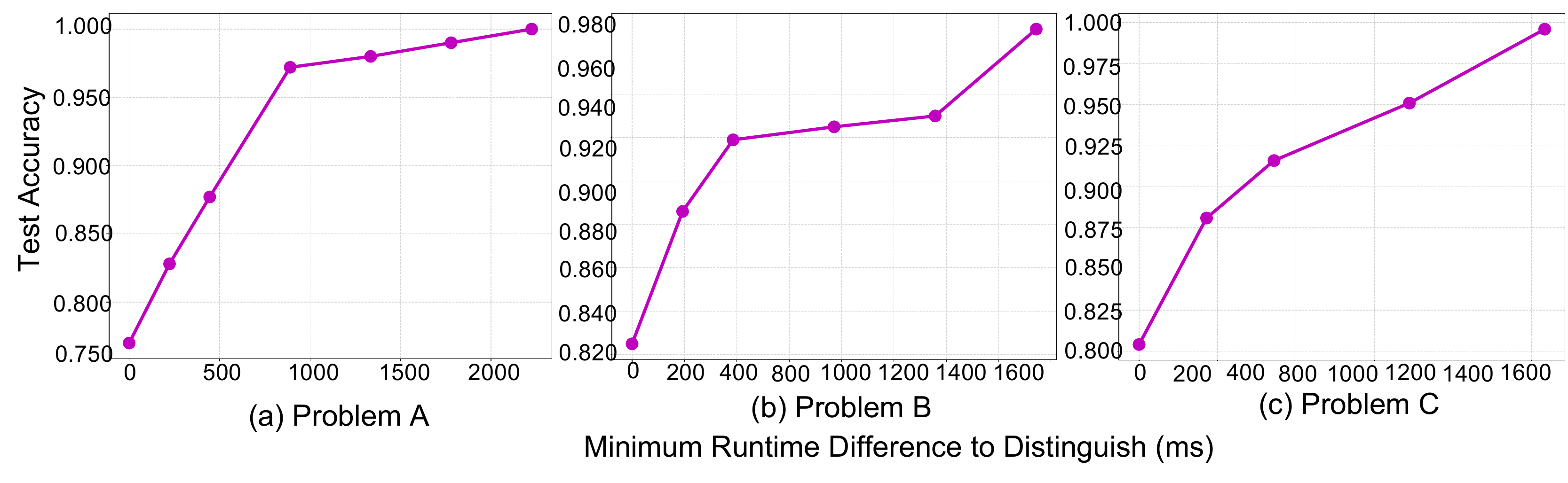}
\caption{Studying the sensitivity of the proposed approach.}
\label{fig:SENS}
\vspace{-0.2in}
\end{figure}
When comparing the execution times, the small differences are less significant than the large differences, e.g., a 1-millisecond difference is treated the same as that with 4000. To evaluate how sensitive the model predictions are to the variation in the submissions' execution times, we sort the evaluation sets and record accuracy for pairs with a difference beyond a certain threshold. 

Figure \ref{fig:SENS} shows the results for models trained on problems A, B, and C. With these three problems, we can observe that the accuracy of the prediction task consistently improves with the increase in the minimum difference that the model needs to resolve. Further investigation uncovers that a massive difference in execution time for source code typically comes from having either loop constructs (e.g., for, while) or significantly longer code. Hence, it becomes easier for the model to spot discriminatory structure in the source code when source code variants differ significantly in execution times. The execution times of all the problems in this dataset are reasonably close. These results indicate that as we move toward problems with a more significant difference in execution time between versions, the model will perform better.

\begin{figure}[t]
\captionsetup{font=small,skip=0pt}
  \centering
\includegraphics[width=\columnwidth]{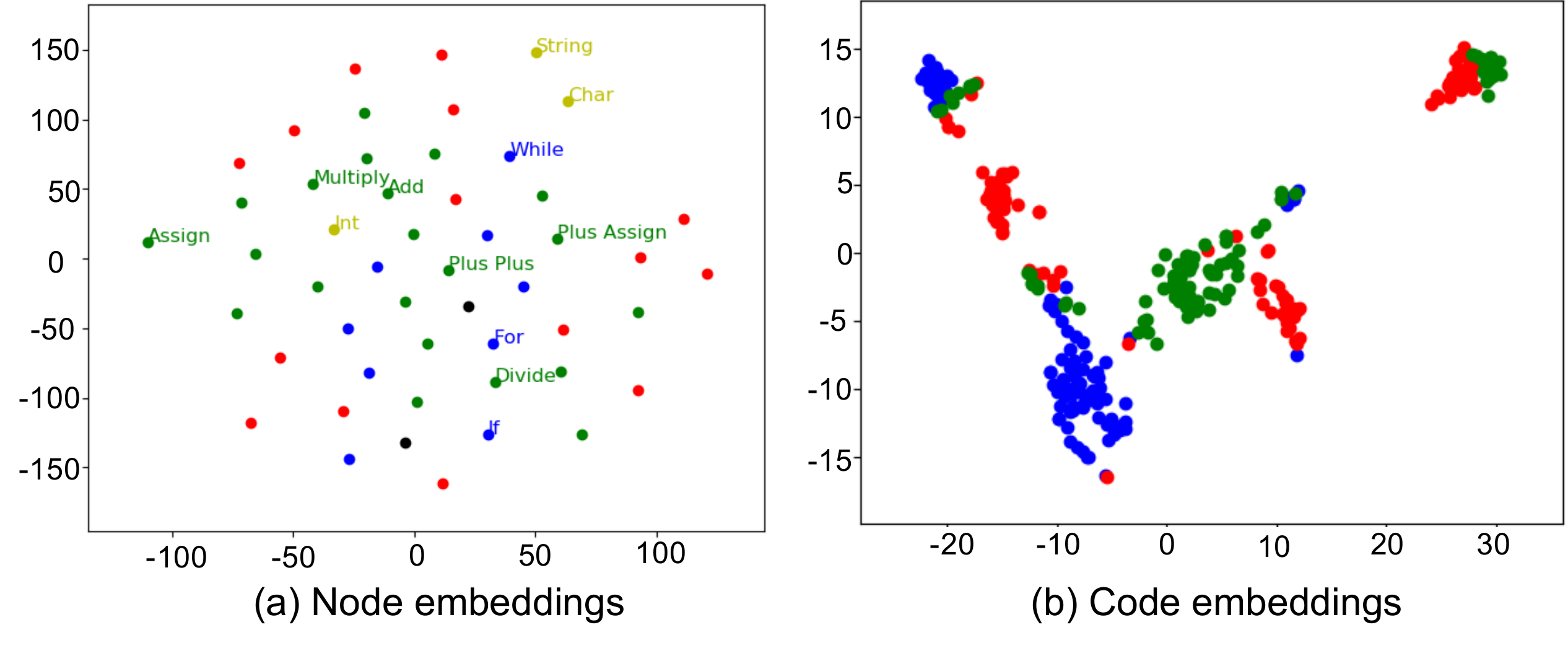}
\caption{Visualization of the learned representations of nodes and ASTs obtained using t-SNE. (a) Two-Dimensional Representation of the node embeddings. Green are operations, red are other expressions, blue are statements, yellow are literal values, and black are support nodes. (b) Two-Dimensional Representation of the AST latent representations. Each color corresponds to one of the problem sets.}
\label{fig:EMBED}
\vspace{-0.2in}
\end{figure}
\subsection{Visualizing Learned Representations}
We initialize nodes with random embedding vectors, and the model subsequently learns representations of each node from the data. To evaluate the effectiveness of the learning process, we map the embeddings down from the $\lambda = 120$ dimensional space to a two-dimensional space and plot in Figure \ref{fig:EMBED}. For the low-dimensional projection, we leverage the unsupervised, non-linear t-Distributed Stochastic Neighbor Embedding (t-SNE) technique~\cite{maaten2008visualizing} that is primarily used for data exploration and visualizing high-dimensional data. 

\noindent\textbf{Node representations: }
Near nodes in t-SNE have similar representations, nodes with equal value along one axis have some similarities, and unequal nodes separated across both axes are significantly different, hence they should have differing representations. Figure~\ref{fig:EMBED}(a) shows that the tree-LSTM model discovers that the string and char literal representations are closely related. 
The model learns that \texttt{plus plus} and \texttt{plus assign} operators are close in nature, hence group them closely. Since the \texttt{for} and \texttt{while} representations share similar values along a single axis, it indicates that the model can encode their similarities into the representations while still capturing their differences.

\noindent\textbf{Code representations: }
Similarly, we use our model to generate code embeddings for three different problems with $100$ submissions each and project those to two-dimensional representations (red, blue, and green). In Figure~\ref{fig:EMBED}(b), we observe that the model can create distinctly different representations from other problems.
We also observe that problems represented with red and green often have grouped clusters indicating more similarity than to the problem in blue.

%% file: future.tex
Figure~\ref{fig:boxplot} shows that the predictive model based on tree-LSTM representations can achieve up to 84\% accuracy for a model built on an individual problem \texttt{E} and 73\% accuracy for model built using a mixture of submissions from hundreds of problems. While the former shows the applicability of the proposed method in studying performance evolution of the same problem over time, the latter demonstrates that in saving time and effort for application developers a majority of the time. 
\change{Once deployed in a continuous learning environment, this framework can improve the model's accuracy by adding new observations based on the nightly tests performed in most production-based software development environments.}

Section~\ref{sec:sensitivity} discusses that it is easier for the model to resolve a higher threshold (large increase in execution time to be considered poor performance) than a smaller one. A developer can select our model through a threshold to prefer true positive cases over false negatives (or vice versa). E.g., with a runtime difference in $1$ second between two applications, the model's accuracy reaches close to 100\%. 
Also, we observe that even though the random initial embeddings are useful, using other sophisticated methods such as Deepwalk~\cite{Perozzi2014DeepWalkOL} to generate initial node representations may help the models to train faster and perform better. Further, these node representations could be made transferable between tasks in analyzing an AST.
Additionally, adding more information, such as the data flow graph, can help the model learn to distinguish problems at a higher level.

%% file: related.tex
The area of software engineering has seen a surge in applying LSTM and its variants in generating code, code summation using natural language~\cite{shido2019automatic}, bug detection~\cite{dam2018deep}, bug report classification~\cite{hanmin2018classifying}, author attribution~\cite{alsulami2017source}, source code retrieval~\cite{wan2019multimodal}, and source code defect prediction~\cite{dam2018deep}.
Machine learning-based program analysis has been studied long in the literature~\cite{lu2012software,canavera2012mining}. Hindle et al.~\cite{hindle2012naturalness} compare programming languages to natural languages and conclude that programs have rich statistical properties. These properties are difficult for a human to capture, but they justify using learning-based approaches to analyze programs.
Others have applied deep learning for performance analysis,~\cite{Cummins2017EndtoEndDL} used sequential LSTMs on normalized source code to predict optimization heuristics. Related work such as~\cite{MouLJZW14,MouLLPJX014} applied a convolutional neural network over the AST to classify the type of problem the source code is trying to solve.~\cite{tai2015improved} performed sentiment analysis for natural language processing using a single layer and directional tree LSTM to build up the representation of the whole tree based upon each node's children, with the leaves being the words in the sentence. Others have looked at using AST for performing static performance analysis~\cite{MengN17}. While using similar methods, including upward and downward passes, their approaches did not include an in-depth learning analysis of the AST in the context of comparative performance prediction.
Neamtiu et al.~\cite{neamtiu2005understanding} have applied deep learning to the program structure information to classify programs. In contrast, our approach, albeit similar, solves a different problem of attributing the change in performance to code structure changes. 

%% file: conclusions.tex
This work aims at designing a generic and accurate method for performance modeling that can provide insight into how a change in the program impacts the performance (positive or negative). We implement a tree-structured deep-neural network architecture to represent the code as a vector to perform the comparison. We then compare two such vector representations from two applications using a feed-forward neural network. This framework can inform a developer about how changes made to their code will affect the performance without necessarily needing to be trained on the specific problem. We trained and evaluated our models on both individual problems and a combined dataset of hundreds of them. Such information available during coding can significantly reduce the need for running applications every time a code change is made and assist in performance-aware application development.